\pdfoutput=1

\documentclass[11pt]{article}

\usepackage[final]{acl}

\usepackage{times}
\usepackage{latexsym}

\usepackage[T1]{fontenc}

\usepackage[utf8]{inputenc}

\usepackage{microtype}
\usepackage{amsmath}

\usepackage{inconsolata}

\usepackage{graphicx}
\usepackage{booktabs}
\usepackage{bbding}
\newcommand{\method}{\textsc{PreAlign}\ }
\newcommand{\methodend}{\textsc{PreAlign}}

\newcommand\blfootnote[1]{%
\begingroup
\renewcommand\thefootnote{}\footnote{#1}%
\addtocounter{footnote}{-1}%
\endgroup
}

\title{\textsc{PreAlign}: Boosting Cross-Lingual Transfer by \\ Early Establishment of Multilingual Alignment}

\author{Jiahuan Li$^\clubsuit$, Shujian Huang$^{\dagger\clubsuit}$, Aarron Ching$^\spadesuit$, Xinyu Dai$^\clubsuit$ \and  Jiajun Chen$^\clubsuit$\\
        $^\clubsuit$National Key Laboratory for Novel Software Technology, Nanjing University, China \\
        $^\spadesuit$Independent Researcher \\
        \texttt{lijh@smail.nju.edu.cn, \{huangsj,daixinyu, chenjj\}@nju.edu.cn}}

\begin{document}
\maketitle
\blfootnote{$\dagger$ The Corresponding author.}
\begin{abstract}
Large language models demonstrate reasonable multilingual abilities, despite predominantly English-centric pretraining.
However, the spontaneous multilingual alignment in these models is shown to be weak, leading to unsatisfactory cross-lingual transfer and knowledge sharing.
Previous works attempt to address this issue by explicitly injecting multilingual alignment information during or after pretraining. Thus for the early stage in pretraining, the alignment is weak for sharing information or knowledge across languages.
In this paper, we propose \methodend, a framework that establishes multilingual alignment prior to language model pretraining.
\method injects multilingual alignment by initializing the model to generate similar representations of aligned words and preserves this alignment using a code-switching strategy during pretraining.
Extensive experiments in a synthetic English to English-Clone setting demonstrate that \method significantly outperforms standard multilingual joint training in language modeling, zero-shot cross-lingual transfer, and cross-lingual knowledge application.
Further experiments in real-world scenarios further validate \methodend's effectiveness across various languages and model sizes. \footnote{The code of this paper is available at \url{https://github.com/Saltychtao/PreAlign}}

\end{abstract}

\section{Introduction}

Large language models~\citep{brown2020language,touvron2023llama,touvron2023llama2} have drastically changed the research paradigm of multilingual language processing. Despite being trained on mainly English texts, they still exhibit reasonable ability for other languages~\citep{touvron2023llama,touvron2023llama2, wang2024seaeval}, and have established multilingual alignment to some extent~\citep{devlin-etal-2019-bert, conneau2019cross,lin-etal-2022-shot}.  However, researchers~\citep{wang2024seaeval,Gao2024MultilingualPA,zhang-etal-2023-dont,qi-etal-2023-cross} have found the spontaneous alignment between languages in these model is still relatively weak, leading to weak cross-lingual factual knowledge retrieval~\citep{wang2024seaeval,Gao2024MultilingualPA} and inconsistency behaviors given the same input~\citep{qi-etal-2023-cross,zhang-etal-2023-dont}.

A handful of works~\citep{reimers-gurevych-2020-making,Cao2020Multilingual,wu-dredze-2020-explicit,chaudhary2020dictmlm,yang-etal-2021-bilingual,tang2022alignmlm,feng-etal-2022-language,Gao2024MultilingualPA} try to mitigate the problem by explicitly injecting alignment information using existing supervision data. They either construct cross-lingual prediction tasks~\citep{chaudhary2020dictmlm,yang-etal-2021-bilingual} or train models to produce similar representations of aligned words or sentences~\citep{tang2022alignmlm,wu-dredze-2020-explicit,reimers-gurevych-2020-making}. However, the improvements are somewhat mixed and the establishment of multilingual alignment requires a long training process either \textit{during} or \textit{after} pretraining~\citep{dufter-schutze-2020-identifying}, which prevents the model from effectively performing cross-lingual transfer at earlier stage in pretraining. 

\begin{figure}
    \centering
    \includegraphics[width=1.0\linewidth]{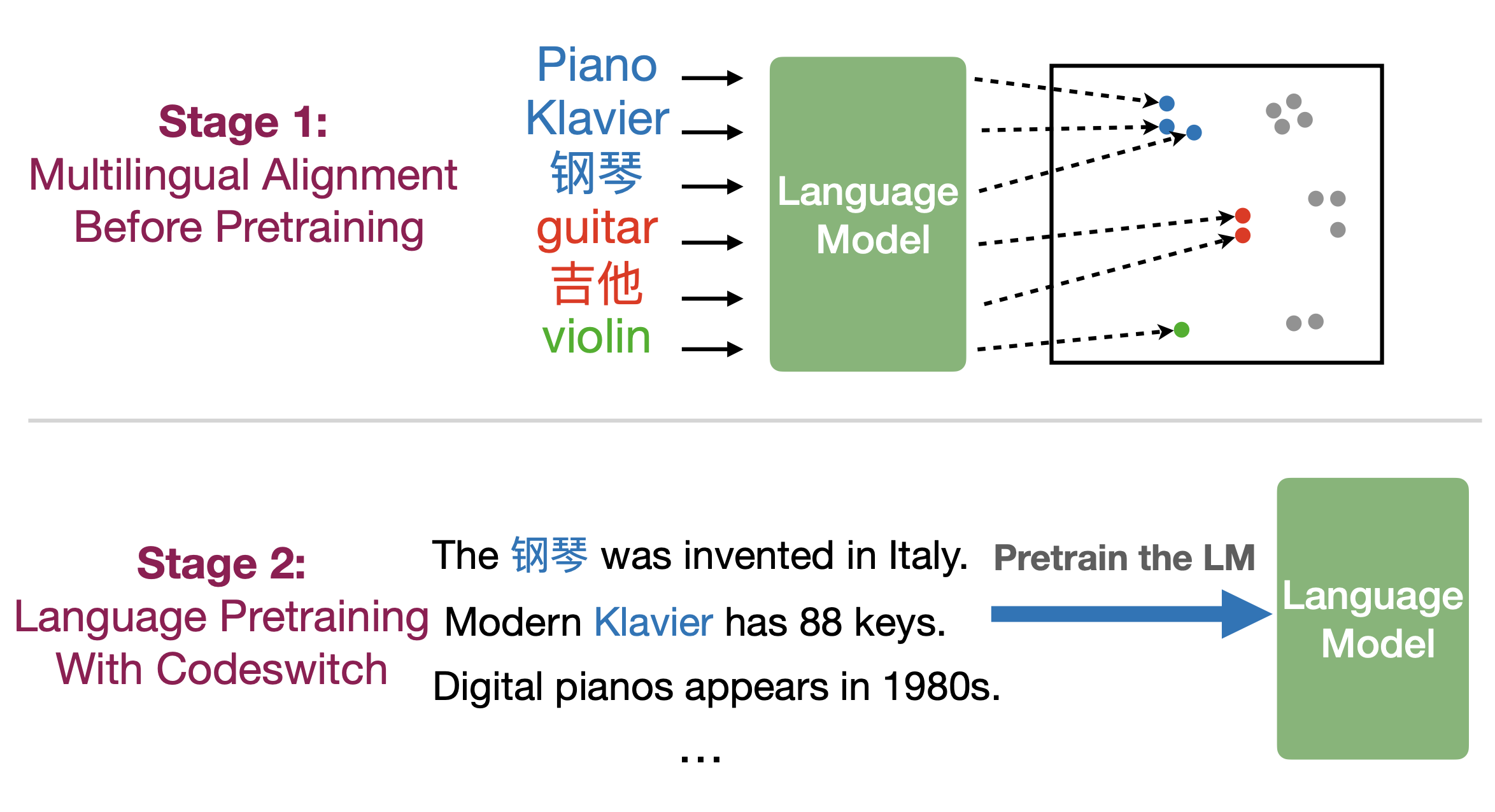}
    \caption{The illustration of \methodend. Words in blue, red and green represent translations of \textit{piano, guitar} and \textit{violin}, respectively.}
    \label{fig:overall_arch}
\end{figure}

In this paper, we introduce \methodend, a framework designed to enhance the alignment of pretrained language models. \method differs from prior methods by integrating the multilingual alignment information \textit{before} extensive language pretraining and maintaining it throughout the pretraining process. This proactive alignment effectively enhances the learning of cross-lingual knowledge in the pretraining corpus, therefore advancing cross-lingual transfer. Therefore, model's proficiency in target languages at early training stage is enhanced, leading to improvement of model's ability to acquire knowledge at that stage.


More specifically, before large-scale language pretraining, \method first collects multilingual translation pairs between English and languages to be transferred, and inject this information into the model by pre-pretraining it to produce similar representations of aligned pairs. In order to maintain the established multilingual alignment across the pretraining phase, we propose an input-only codeswitching strategy, which only substitutes words in the input text to its aligned words, and optimizes model using language modeling objective. The illustration of \method is shown in Figure~\ref{fig:overall_arch}.

We firstly conduct experiments on an English to English-Clone setting~\citep{K2020Cross-Lingual, dufter-schutze-2020-identifying,schäfer2024language}, where English-clone is a synthetic language that shares identical grammar and vocabulary distribution with English, but has no vocab overlap. This allows us to study cross-lingual transfer on a more controlled environment.
Experiments demonstrate that \method improves language ability on English-Clone by strengthening the cross-lingual transfer of knowledge and abilities from English. Further analysis shows that the early established multilingual alignment can be kept throughout large-scale language pretraining and generalize to other unaligned words.
Experiments on real-world settings (including Chinese, German, Arabic and Russian) validate the effectiveness of \method across different languages and model scales.



\section{Related Work}

\subsection{Understanding Cross-lingual Ability of Pretrained language models}

Many works attempt to analyze the cross-lingual ability of LLMs. \citet{dufter-schutze-2020-identifying,conneau-etal-2020-emerging} try to explain factors that contributes to spontaneous multilingual alignment developed in pretrained language models, including under-parameterization, shared model architectures and pivot words across languages. Other works investigate the working mechanism of multilingual representations. \citet{wendler2024llamas} find that English-centric models works on a concept space that is close to English when processing other languages. \citet{gaschi-etal-2023-exploring,hammerl-etal-2024-understanding} discuss the relationship between multilingual alignment and cross-lingual transfer. Recently, \citet{Gao2024MultilingualPA,qi-etal-2023-cross} analyze multilingual knowledge alignment in existing LLMs, and find that multilingual training and instruction tuning can only lead to shallow alignment, i.e. LLMs can achieve similar task performances and consistent responses across languages, yet cannot apply knowledge across languages.

Our paper differs from theirs in that we focus on improving models' cross-lingual ability and successfully unlocks the ability of cross-lingual knowledge transferring.

\subsection{Enhancing Cross-lingual Ability of Pretrained Language Models} Other studies also seek to enhance the cross-lingual capabilities of pretrained language models. These typically utilize explicit alignment signals, such as parallel sentences and dictionaries. They can be categorized based on when the alignment occurs: during pretraining or post-pretraining.

On the first category, \citet{Yang_Ma_Zhang_Wu_Li_Zhou_2020,chaudhary2020dictmlm} perform codeswitching on the monolingual data to make model better capture cross-lingual relation and dependency. \citet{hu-etal-2021-explicit} train the model to produce consistent word alignment matrices between source and target language and similar representations for parallel sentences. \citet{chi-etal-2022-xlm} explores multilingual replaced token detection and translation replaced token detection task. \citet{tang2022alignmlm} further maximize the cosine similarity of aligned word embeddings to explicitly inject multilingual alignment. 

On the second category, researchers enhance the multilingual alignment after pretraining. Earlier works either optimizes pretrained models to produce similar representations for parallel sentences~\citep{reimers-gurevych-2020-making,pan-etal-2021-multilingual,feng-etal-2022-language} or parallel words~\citep{Cao2020Multilingual,wu-dredze-2020-explicit}. Recent works on large language models typically train the model to produce consistent responses~\citep{She2024MAPOAM} or performing cross-lingual instruction-following tasks~\citep{zhu2024question,zhu2024power}.

\method differs from all above works in that it establishes multilingual alignment before language pretraining, therefore facilitating the cross-lingual transfer at early pretraining stage.

\section{The PreAlign Method}
In this section, we present \methodend, a simple and effective framework that advances the establishment of multilingual alignment before language pretraining. 

\subsection{Injecting Multilingual Alignment before Language Pretraining}

\method aims to inject multilingual alignment information before large-scale language model pretraining, which facilitates cross-lingual transfer as soon as possible. This involves two stages: {collection of multilingual alignment table} and {alignment injection via contrastive learning}. 

\paragraph{Collection of multilingual alignment table} The collection of multilingual alignment table can either leverage the off-the-shelf multilingual dictionaries such as MUSE, or rely on machine translation models. In this paper, we take the second method: we first extract from an English monolingual corpus $\mathcal{D}$ the collections of all unique words $\mathcal{W} = \{w\}_i^N$, where $N$ is the number of unique words. For each word $w$, we translate it to all considered target languages, and denote the translation results as $T(w)$. We collect diverse translations for each word using GPT-4.  More details can be found in Appendix~\ref{sec:experimental_details}.


\paragraph{Alignment injection via contrastive learning} After the multilingual alignment table is collected, \method initializes models' parameters using a contrastive alignment objective, which optimizes the model to produce similar representations for aligned words.  Specifically, given an English word $w_i$ and its available translations across other languages $T(w_i)$, \method firstly obtains representations of each layer for each $w \in T(w_i)$:
\begin{equation}
    h_{w}^l = \text{MeanPool}(f(w,l))
\end{equation}

\noindent where $l=0,1,\cdots,L,L+1$; $f(w,l)$ for $1 \le l \le L$ denotes the $l$-th Transformer layer representations of the model's encoding of $w$; $f(w,0)$ and $f(w,L+1)$ denotes the word embedding and output embedding of $w$, respectively. Note that since $w$ could be tokenized to multiple subwords, \method aggregates them into a single representation using mean-pooling operator. 

\method then leverages a contrastive learning objective~\citep{khosla2021supervised} to establish alignments between words in different languages:

\begin{equation}
\mathcal{L}^l_{\text{align}} = \sum_{\substack{w_j \in \mathcal{W} \\ w_i \in T(w_j)}} \text{log} \frac{\exp(\ d(h^l_{w_i}, h^l_{w_j})/\tau)}
{\sum_{w_k \in \mathcal{B}} \exp(\ d(h^l_{w_j}, h^l_{w_k})/\tau)} \label{constrastive}
\end{equation}


\noindent where $\mathcal{B}$ is the set of all words in current mini-batch, $\tau$ is the temperature parameter. $cos(\cdot, \cdot)$ is the cosine similarity function. The final learning objective is the sum of contrastive loss of all layers:

\begin{equation}
    \mathcal{L}_{\text{align}} = \sum_{l=0}^{L+1} \mathcal{L}_{\text{align}}^l
\end{equation}

To prevent the initialization from being trapped in a local minima that is not suitable for the subsequent language modeling, we also add an auxiliary language modeling loss beside the contrastive objective in practice~\footnote{The training data for $\mathcal{L}_\text{align}$ and $\mathcal{L}_\text{LM}$ are independently sampled in each mini-batch.}:

\begin{equation}
    \mathcal{L}_{\text{joint}} = \mathcal{L}_{\text{align}} + \mathcal{L}_{\text{LM}}
\end{equation}

\noindent Note that, the $\mathcal{L}_{LM}$ objective in the pre-alignment stage only serves to regularize the optimization process, rather than performing large-scale pre-training. In practice, this stage only consumes 5\% pretraining data.

\subsection{Maintaining Multilingual Alignment via Input-only Codeswitching}
\method injects multilingual alignment information before language pretraining. However, it is possible that this information could be quickly forgotten if not continuously reinforced. Inspired by prior research ~\citep{chaudhary2020dictmlm,yang-etal-2021-bilingual} demonstrating that codeswitching effectively promotes multilingual alignment, we propose using the codeswitching technique to sustain this alignment throughout the pretraining process.

\begin{figure}
    \centering
    \includegraphics[width=0.9\linewidth]{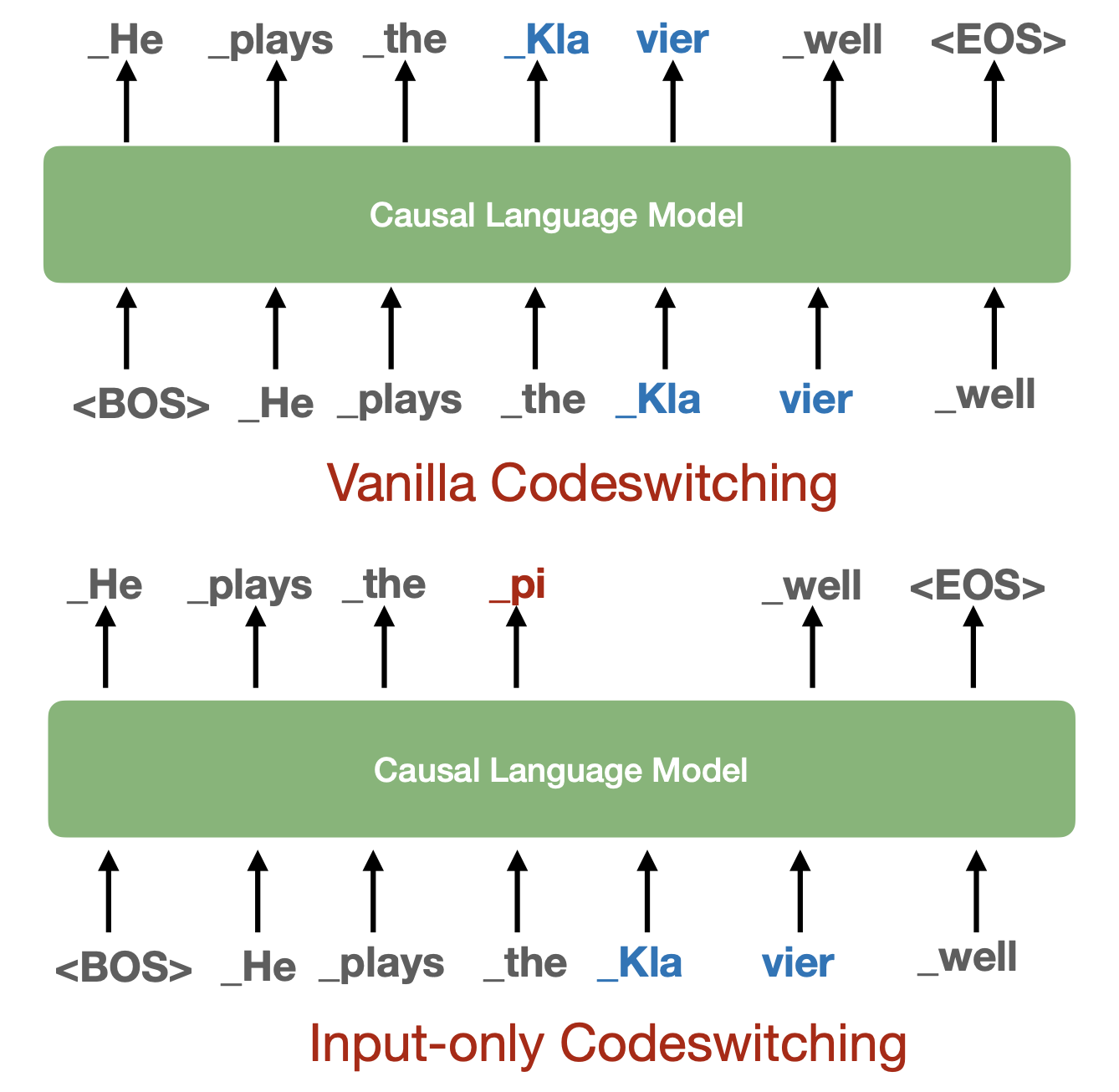}
    \caption{Comparison between vanilla codeswitching and the proposed input-only codeswitching. The original English sentence is \textit{He plays the piano well}, and \textit{Klavier} is the German translation of \textit{piano}.}
    \label{fig:input_only}
\end{figure}

Originally, codeswitching was applied to both the input sequence and the target tokens in raw data, which can exacerbate the issue of multilingual script mixing in the outputs of decoder-only models. To address this, we propose an input-only codeswitching strategy that affects only the input. The distinction between the traditional codeswitching and our input-only codeswitching is illustrated in Figure~\ref{fig:input_only}.

Formally, given a subword sequence $X_{<i},x_{i}^1, \cdots ,x^{i}_{m},X_{>i}$, where $X_{<i}$ and $X_{>i}$ are the subword sequences before and after the $i$-th word, respectively; $x_{i}^1, \cdots ,x_{i}^{m}$ is the subword sequence of the $i$-th words. Suppose the $i$-th word is substituted by $y_{i}^1, \cdots ,y_{i}^{n}$ after codeswitching, then the language modeling objective after the original codeswitching is 
\vspace{-0.1cm}
\begin{align}
   \mathcal{L}_{\text{vanilla\_CS}} =  p(X_{<i}) & \cdot p(X_{>i} | X_{<i},y_{i}^1, \cdots ,y_{i}^{n})  \notag  \\
     &  \cdot p(y_i^{1}|X_{<i}) \notag \\ & \cdot \prod_{j=2}^n p(y_i^{j}|X_{<i},y_{i}^1, \cdots ,y_{i}^{j-1}) \label{eqn:original_codeswitch}
\end{align}

\noindent In Equation \ref{eqn:original_codeswitch}, the item $p(y_i^1|X_{<i})$ requires the model to generate words in one language given prefixes in another language. To mitigate this, our input-only codeswitching modifies the objective to be
\begin{align}
    \mathcal{L}_{\text{input\_only\_CS}} =     p(X_{<i}) & \cdot p(X_{>i} | X_{<i}, y_{i}^1, \cdots ,y_{i}^{n}) \notag \\ 
    & \cdot p(x_i^{1}|X_{<i}).  \label{eqn:input_only_codeswitch}
\end{align}

\noindent Equation \ref{eqn:input_only_codeswitch} omits the prediction objective of subwords in the word after codeswitching ($p(y_i^{1}|X_{<i})$), therefore preventing the generation results contain scripts from other languages. In this paper, we use a codeswitching ratio of 5\%.




\section{Evaluation of Cross-Lingual Transfer}
To investigate the cross-lingual transfer effects that is close to situations in current LLMs, we design the evaluation in an English-dominated setting, where most of the pretraining data is English. For the examined target language, the amount of pretraining data is much less than English. Intuitively, the language ability of the target language will be much weaker than English. However, it is still interesting to know to what extent the language abilities and knowledge could transfer from English to the target language.

\subsection{Languages}


\begin{figure}
    \centering
    \includegraphics[width=\linewidth]{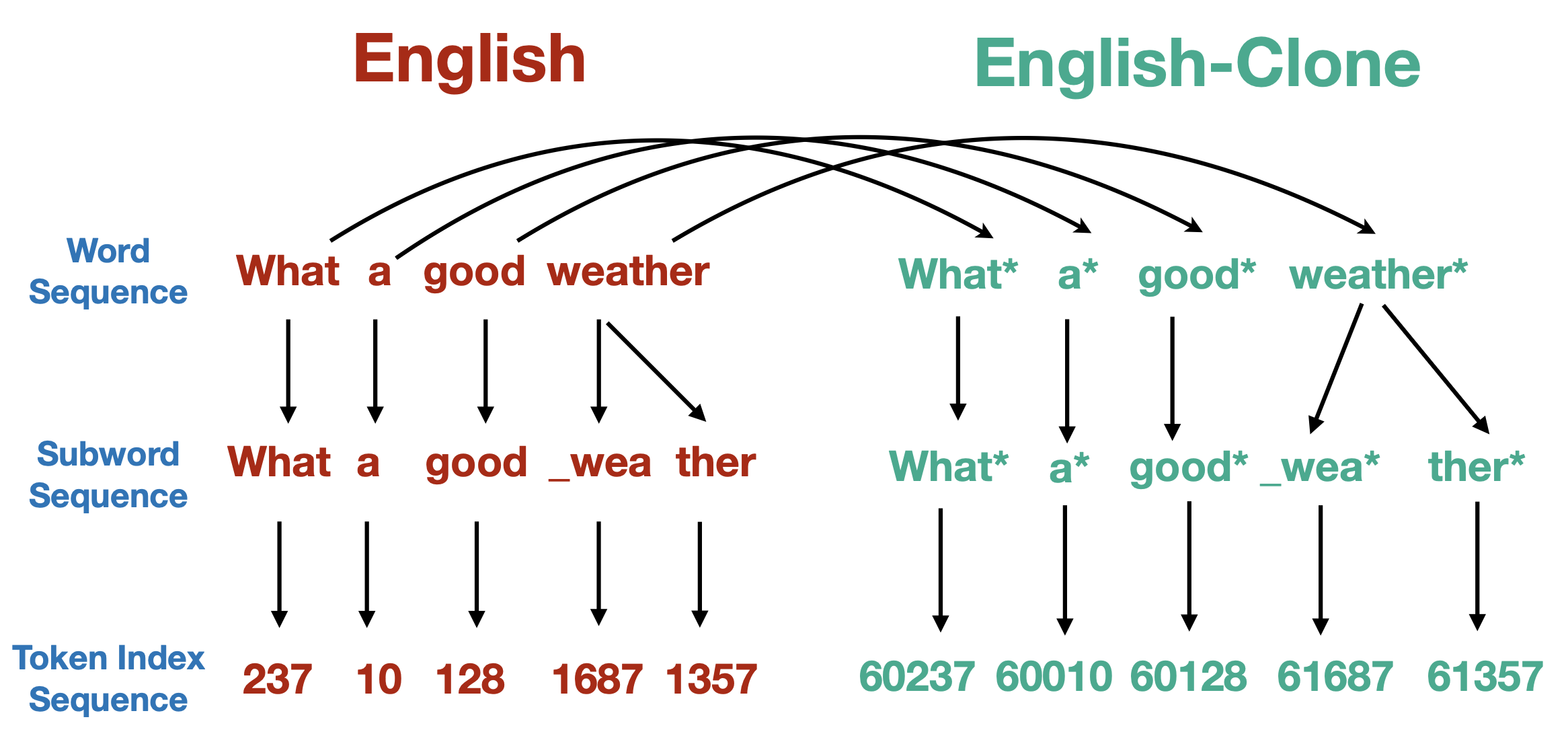}
    \caption{Illustration of the creation of English-Clone.}
    \label{fig:en_clone}
\end{figure}

\subsubsection{Synthetic Language: En-Clone}
\label{sec:4.1}
We construct a synthetic language called En-Clone, by cloning all English tokens by a one-to-one mapping. En-Clone shares the same linguistic properties with English, such as vocabulary distribution, subword segmentation, grammar and syntax, yet they have no word overlapping. See Figure~\ref{fig:en_clone} for an illustration. 

This synthetic setting provides many benefits. Firstly, the English to En-Clone setting arguably forms the easiest setting for testing the cross-lingual transferring ability of LLMs, since it does not involve the discrepancy of word ordering and possibly complex one-to-many/many-to-one alignments between real-world languages. Therefore, this setting can serve as a sanity-check for cross lingual transferring methods.

Secondly, since the golden alignment between English and En-Clone is trivial to get, we can easily achieve \textit{perfect} alignment at the initialization stage by setting the input and output embedding of aligned tokens to be identical. In this way, hidden states of all intermediate layers would also be identical. This provides us a chance to analyze the upper-bound performance of our method.


\subsubsection{Real-World Languages}
We also experiment with real-world languages to examine the effect of \method in more complex situations. We select Chinese, Russian, German and Arabic as our target languages, which spans four different language families, serving as good representatives of world languages. Note that in this case, the alignment is established for 4 target languages at the same time.

\subsection{Evaluation Metrics}
The cross-lingual transfer effects are evaluated in the following 3 aspects:

\paragraph{Target Language Modeling (LM)} The first evaluation metric is the language modeling performance (perplexity) of the target language. Given the same amount of target language data, this can reflect the extent of language ability transferred from English to the target language.

\paragraph{Zero-shot Cross-lingual Transfer (ZS-CLT)} Another common way to evaluate model's cross-lingual ability is zero-shot cross-lingual transfer, where we finetune models with training data of a given task in the source language, and test model's ability on the same task in target languages. We use the commonly-used XNLI~\citep{conneau-etal-2018-xnli} dataset for ZS-CLT evaluation.

\paragraph{Cross-lingual Knowledge Application (CLKA)} 

Large language models acquire extensive world knowledge from their pretraining corpora, which might be described in different languages. 
It is ideal for LLMs to learn knowledge from texts in one language and apply it across other languages.

In order to evaluate models' ability to perform such cross-lingual knowledge application, we propose a setting where the model is trained with certain English texts describing synthetic knowledge, and test the injected knowledge in the target language. Each synthetic knowledge is a triplet like (subject, relation, object), where relations are extracted from WikiData~\citep{wikidata}, and subjects and objects are artificial entities. We assess the model's knowledge retention by comparing the likelihood of different statements, including one correct statement and three distractors generated by randomly substituting named-entities for the original object in the knowledge statement. See Appendix~\ref{sec:experimental_details} for examples of synthesized knowledge.


\subsection{Experiment Settings in General}

\paragraph{Pretraining Dataset} We adopt CulturaX~\citep{nguyen2023culturax} as the pretraining dataset. CulturaX is a multilingual pretraining corpus that has been rigorously cleaned. 
For English, we randomly select 10 billion tokens from CulturaX as the pretraining data. For each language to be transferred to (Zh, De, Ru and Ar in the real-world setting, and En-Clone in the synthetic setting), we randomly select 100 million tokens, which is 1\% of the data in English. 

\paragraph{Model Configuration} We adopt the GPT-2 style Transformer architecture for our model. As the defaulting setting, our model contains 12 Transformer layers with a hidden dimension of 1024. The number of total non-embedding parameters is about 150 million. We use AdamW~\citep{kingma2017adam} optimizer with a global batch size of about 1 million tokens. The learning rate is decayed from $3e-4$ to $3e-5$ following a cosine scheduler. 

\paragraph{Baselines} We compare \method's performance with the following methods:

\begin{itemize}
    \item Joint Training, where we pretrain the model on the mix of 10 billion English tokens and 100 million tokens in the target language.
    \item Only Target, where we only pretrain the model on 100 million tokens in the target language.
    \item Full Target, where we pretrain the model on 10 billion tokens in the target language. This can serve as an upper-bound performance for the target language.
\end{itemize}

\begin{table*}[!ht]
\centering
\footnotesize
\begin{tabular}{ccccccccc}
\toprule
\multicolumn{1}{l}{} & \multicolumn{2}{c}{\#Tokens} & \multicolumn{2}{c}{LM (ppl. $\downarrow$)} & \multicolumn{2}{c}{ZS-CLT (acc. $\uparrow$)} & \multicolumn{1}{c}{CLKA (acc. $\uparrow$)} \\ 
\multicolumn{1}{l}{} & En  & En-Clone & En   & En-Clone & En   & En-Clone &  En-Clone \\ \midrule
Only Target        & -   & 0.1B     &   -   &    47.2      &   -   &    -      &   -    \\
Joint Training       & 10B & 0.1B     & 16.1 & 21.6     & 79.8 & 74.9     & 26.5     \\
\method             & 10B & 0.1B     & \textbf{15.9} & \textbf{16.5}     & \textbf{80.1} & \textbf{79.3}     & \textbf{90.3}     \\ \midrule
Full Target        & -   & 10B      &   -   &    16.2      &   -   &    -      &   -         \\
\bottomrule
\end{tabular}
\caption{Performance of \method and other methods on language modeling, zero-shot cross-lingual transfer (ZS-CLT) and cross-lingual knowledge application (CLKA). }
\label{tab:main_result}
\end{table*}

\section{Experiments on Synthetic Setting}

e start our evaluation on the English to En-Clone setting, which allows us to better control the relationship between the source and target language. 

\subsection{General Results}
\label{sec:4.2}
We present results on LM, ZS-CLT and CLKA in Table~\ref{tab:main_result}. 

\paragraph{Joint Training achieves spontaneous multilingual transfer to some extent.}  Table~\ref{tab:main_result} shows that compared to Only-Target, Joint training achieves notable improvements on LM despite there are neither parallel signal or pivot words between English and English-clone. Surprisingly, the model could successfully transfer the ability to perform NLI task from English (79.8) to English-Clone (74.9).  However, this transfer does not work well on CLKA, which is consistent with previous findings~\citep{Gao2024MultilingualPA} that cross-lingual knowledge transfer is hard to achieve by multilingual pretraining.

\paragraph{\method improves over Joint Training on all evaluation tasks.} We can also see that \method significantly outperforms Joint Training on all three evaluation metrics. On the LM evaluation, \method even achieves performance comparable to Full-Target, using only 1\% data.  For ZS-CLT, the performance gap between two languages are narrowed. For the CLKA the accuracy is greatly improved (from 27.7 to 64.6).  All the results demonstrate the effectiveness of \method for facilitating cross-lingual transfer. It is worth noticing that the performance of English are also improved, suggesting the learning of English-Clone also helps English as well.

\paragraph{\method outperforms methods that establish alignment during and post pretraining.}
We experiment with performing contrastive alignment during and post the pretraining process~\citep{wu-dredze-2020-explicit}, and compared them to \method in Table~\ref{tab:alignment_stage}. For the post-pretraining alignment, we add an additional LM loss to reduce catastrophic forgetting. It can be seen that on-the-fly alignment can degrade model's performance, which we hypothesize that excessively optimizing the model on the limited size of word-level multilingual alignment during pretraining might have a negative impact on language ability. For post-pretraining alignment, we can observe a improvement over Joint Training on LM and ZS-CLT, but nearly no effect on CLKA. However, PreAlign outperforms both on-the-fly alignment and post alignment on all three evaluation protocols.

\begin{table}[t]
\centering
\footnotesize
\begin{tabular}{lccc}
\toprule
                                      & \multicolumn{1}{l}{\textbf{LM}} & \multicolumn{1}{l}{\textbf{ZS-CLT}} & \multicolumn{1}{l}{\textbf{CLKA}} \\ \midrule
Joint Training                        & 21.6                            & 74.9                                & 26.5                              \\
On-the-fly alignment & 22.1                            & 74.3                                & 26.5                              \\
Post alignment       & 19.7                            & 75.5                                & 28.4                              \\
PreAlign                              & \textbf{16.5}                   & \textbf{79.3}                       & \textbf{90.3}      \\ \bottomrule              
\end{tabular}
\caption{Comparison of performing contrastive alignment at different stage. On-the-fly alignment: performing alignment during the pretraining. Post alignment: performing alignment when the pretraining is done.}
\label{tab:alignment_stage}
\end{table}

\subsection{In-Depth Investigation for CLKA}
\label{sec:4.3}
\begin{figure*}[t]
\centering
\includegraphics[width=0.98\linewidth]{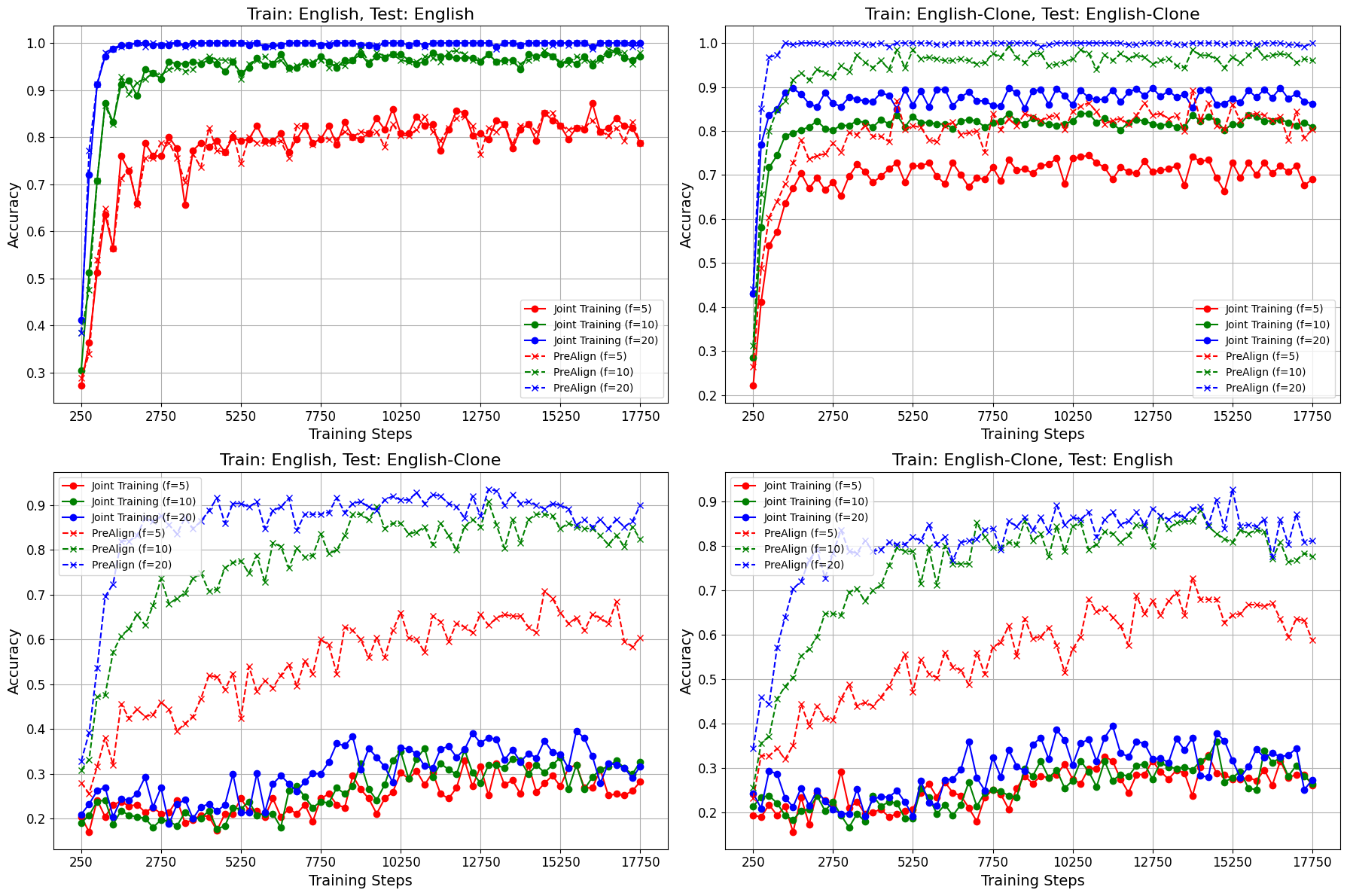}
\caption{Knowledge application accuracy at each training period of different models. $f$ indicates the frequency of the test knowledge.}
\label{fig:knowledge_transfer}
\end{figure*}

As the performance for CLKA varies for different methods, we further examine the learning dynamics of CLKA. We segment the pretraining process into shorter periods, each consisting of 250 training steps, and evaluate the CLKA accuracy for each period. More specifically, for each period, knowledge of different frequence are provided to the model, and 
assessment occurs immediately after each  period using the corresponding model checkpoint. For comparison, we evaluate all four combinations of languages for training and testing the knowledge. 
Figure~\ref{fig:knowledge_transfer} shows the results.

\paragraph{Knowledge learning ability correlates with language ability.} We can see from the top-left of Figure~\ref{fig:knowledge_transfer}, where we test English knowledge in English language, models' knowledge completion accuracy after each learning period rapidly grows as the pretraining goes on. This indicates that the models' ability to acquire knowledge correlates with their language modeling ability. The final performance also correlates with the knowledge frequency in the learning period as expected.

\paragraph{Early cross-lingual transfer enhance target language ability, facilitating knowledge learning.} In the top-right of Figure~\ref{fig:knowledge_transfer} where we test English-Clone knowledge in English-clone language, we observe a similar trend as the top-left figure. However, the growing rate of \method is higher compared to Joint Training especially when frequency of knowledge is low, thanks to better transfer of language ability from English to English-Clone.

\paragraph{\method unlocks CLKA.} From the bottom two figures in Figure~\ref{fig:knowledge_transfer}, we can see the CLKA ability of Joint Training is greatly weaker than \methodend, close to the random guessing performance. This renders \method a promising method for learning truly multilingual knowledge alignment.

\subsection{Ablation Study}
\label{sec:4.4}

\begin{table*}[t]
\centering
\small
\begin{tabular}{l|ccc|ccc}
\toprule
                        & Joint Training & Multi-Align Init & Input-only CS & LM (ppl. $\downarrow$)   & ZS-CLT (acc. $\uparrow$) & CLKA  (acc. $\uparrow$)\\ \midrule
\#1                     & \Checkmark              &                                       &               & 21.6 & 74.9 & 26.5 \\
\#2                     & \Checkmark              &                                      & \Checkmark             & 19.7 & 76.1 & 30.2 \\
\#3                     & \Checkmark              & \Checkmark                                   &               & 17.1 & 77.8 & 85.7 \\
\#4                     & \Checkmark              & \Checkmark                                     & \Checkmark             & \textbf{16.5} & \textbf{79.3} & \textbf{90.3} \\ \bottomrule
\end{tabular}

\caption{Ablations of \methodend. Multi-Align Init: using multilingual alignment objective to initialize LM. Input-only CS: the proposed data augmentation method by only codeswitching the input words. All reported performance are evaluated in English-Clone.}
\label{tab:ablation}
\end{table*}

In this section, we present an ablation study of the proposed methods. The results are in Table~\ref{tab:ablation}. 

\paragraph{Solely input-only CS helps LM and ZS-CLT, but not CLKA.} Comparing Line \#1 and Line \#2, we can see that adding input-only CS to the pretraining stage can bring improvements to language modeling and downstream cross-lingual transferring performance, which is consistent with findings in previous works~\citep{chaudhary2020dictmlm,yang-etal-2021-bilingual}. However, the improvement on CLKA is much smaller (27.7 $\to$ 32.6).

\paragraph{Multilingual alignment initialization significantly facilitates CLT, especially CLKA.} By establishing multilingual alignment before language model pretraining, all considered metrics that evaluating cross-lingual transfer are significantly improved (Line \#1 vs. Line \#3 and Line \#2 vs. Line \#4). Notably, this brings a much better CLKA performance, highlighting the importance of early multilingual alignment for knowledge transferring.

\paragraph{Combining Multi-Align Init with input-only codeswitching achieves the best performance.} Finally, by comparing Line \#4 vs. Line \#2 and Line \#3, we can see the proposed two strategies all contributes to the good performance that \method achieves.

\paragraph{Input-only codeswitching causes less mixed-script problem.}
We also compare the proposed input-only codeswitching strategy with the vanilla codeswitching strategy in Table~\ref{tab:codeswitch_ratio}, in terms of both English language modeling performance and the ratio that generation results contains En-clone tokens. It can be seen that when the training time codeswitching ratio is to 5\%, adopting vanilla codeswitching strategy would result in 4.17\% sentences contains En-clone tokens, which would significantly decrease the generation quality in real-world settings. However, the input-only codeswitching strategy proposed in this paper effectively decrease the ratio to 0.02\%, and achieves better English LM perplexity.

\begin{table}[t]
\centering
\footnotesize
\begin{tabular}{ccc}
\toprule
              & LM   &Mixed-Script Ratio \\ \midrule
Original CS   & 17.1 & 4.17\%            \\
Input-only CS & 16.5 & 0.02\%           \\ \bottomrule
\end{tabular}
\caption{Comparison of the original codeswitching strategy and the proposed input-only codeswitching strategy. Note the mixed-script ratio in the table refers to the portion of random English samples that contains English-clone scripts during inference.}
\label{tab:codeswitch_ratio}
\end{table}

\begin{figure}
    \centering
    \includegraphics[width=\linewidth]{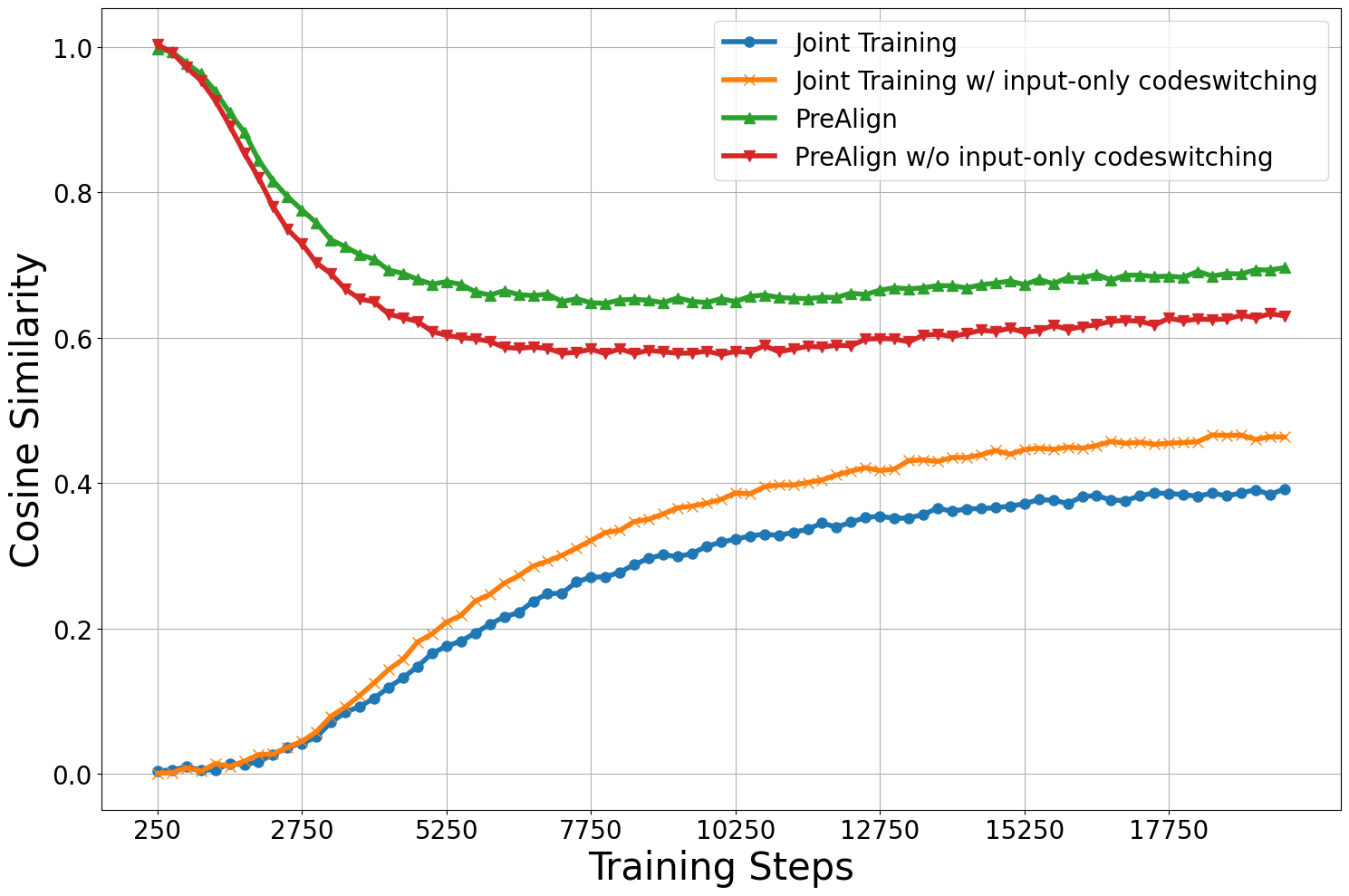}
    \caption{The evolution of word embeddings' cosine similarity between aligned words from different models.}
    \label{fig:cosine_score}
\end{figure}

\begin{table}[t]
\centering
\footnotesize
\begin{tabular}{@{}lccc@{}}
\toprule
\multicolumn{1}{l}{}               & LM            & ZS-CLT          & CLKA          \\ \midrule
\multicolumn{1}{c}{Joint Training} & 21.6          & 74.9          & 26.5          \\ \midrule
\multicolumn{4}{l}{\method}                                                       \\ \midrule
\ \ \ \ \ \ \ \ $\beta=25\%$                          & 17.0          & 78.2          & 80.2          \\
\ \ \ \ \ \ \ \ $\beta=50\%$                           & 16.8          & 78.6          & 83.1          \\
\ \ \ \ \ \ \ \ $\beta=75\%$                          & 16.6          & 78.8          & 88.4          \\
\ \ \ \ \ \ \ \ $\beta=100\%$                           & \textbf{16.5} & \textbf{79.3} & \textbf{90.3} \\ \bottomrule
\end{tabular}
\caption{Performance of \method when using different portion of aligned word pairs. For reference, we also list the performance of Joint Training.}
\label{tab:unseen_portion}
\end{table}

\subsection{Maintaining Multilingual Alignment across Pretraining.}
\label{sec:4.5}

\begin{table*}[t]
\footnotesize
\begin{tabular}{lccccccccccccccc}
\toprule
               & \multicolumn{5}{c}{LM(ppl. $\downarrow$)}                                                       & \multicolumn{5}{c}{ZS-CLT(acc. $\uparrow$)}                                                      & \multicolumn{4}{c}{CLKA(acc. $\uparrow$)} \\ \midrule
               & En            & Zh            & De            & Ar            & Ru            & En            & Zh            & De            & Ar            & Ru            &  Zh  & De & Ar & Ru \\  \midrule
\textbf{150M}           &               &               &               &               &               &               &               &               &               &               &     &     &    &    &    \\ \midrule
Joint Training & 25.7          & 99.7          & 43.5          & 46.9          & 49.8          & \textbf{80.6} & 64.6          & 63.5          & 58.3          & 62.0          &   26.2   &  25.1  & 26.8   &  26.3  \\
\method       & \textbf{25.4} & \textbf{91.1} & \textbf{39.8} & \textbf{40.7} & \textbf{44.6} & \textbf{80.6} & \textbf{69.2} & \textbf{67.5} & \textbf{60.8} & \textbf{65.1} &   \textbf{53.1}   & \textbf{57.2}   &  \textbf{51.6}  & \textbf{55.5}   \\ \midrule
\textbf{400M}           &               &               &               &               &               &               &               &               &               &               &     &     &    &    &    \\ \midrule
Joint Training & 20.3          & 79.8          & 32.5          & 34.8          & 39.6          & 82.3          & 65.8          & 65.3          & 56.9          & 63.7          &     37.8   &  39.5  &  36.1  &  37.7  \\
\method       & \textbf{19.9} & \textbf{75.2} & \textbf{28.3} & \textbf{30.7} & \textbf{33.6} & \textbf{82.4} & \textbf{70.0} & \textbf{69.3} & \textbf{65.6} & \textbf{68.2} &   \textbf{63.8}   &  \textbf{66.5}  &  \textbf{64.7}  & \textbf{63.6}   \\ \midrule
\textbf{1.3B}           &               &               &               &               &               &               &               &               &               &               &     &     &    &    &    \\ \midrule
Joint Training & \textbf{15.8} & 62.2          & 24.0          & 27.7          & 31.2          & \textbf{84.3} & 70.8          & 70.6          & 63.7          & 68.6          &   49.6   & 44.1   & 45.5   & 48.0   \\ 
\method       & 16.1          & \textbf{58.0} & \textbf{23.3} & \textbf{25.3} & \textbf{29.4} & 83.9          & \textbf{74.0} & \textbf{72.9} & \textbf{68.2} & \textbf{71.4} &    \textbf{71.1}   &  \textbf{73.9}  &  \textbf{72.7}  &  \textbf{72.5}  \\ \bottomrule
\end{tabular}
\caption{Performance of Joint Training and \method across different scale of models on language modeling, zero-shot cross-lingual transfer (ZS-CLT) and cross-lingual knowledge application (CLKA).}
\label{tab:real_world}
\end{table*}

In order to understand how the injected multilingual alignment information evolves during pretraining, we compute the similarity of  aligned word embedding at different training period (every 250 training steps). Figure~\ref{fig:cosine_score} illustrates the results. 

Firstly, we can see that despite there are no vocabulary overlap between English and English-clone, the embedding similarity of aligned words still grows during Joint-Training, which is consistent with findings in previous works~\cite{dufter-schutze-2020-identifying}. This indicates the ability of spontaneous establishment of multilingual alignment of language models. Secondly, the aligned similarity score of \method is near perfect as designed, and despite the score decreases at the beginning of pretraining, it maintains to be significantly higher than Joint-Training throughout the pretraining process. Finally, the codeswitching strategy is helpful for both Joint Training and \methodend, as it accelerates the increment of Joint Training's aligning similarity score, and helps slow down the decrement of \methodend's aligning similarity score.

\subsection{Generalization to Unseen Word Pairs}
\label{sec:4.6}


\begin{figure}
    \centering
    \includegraphics[width=0.85\linewidth]{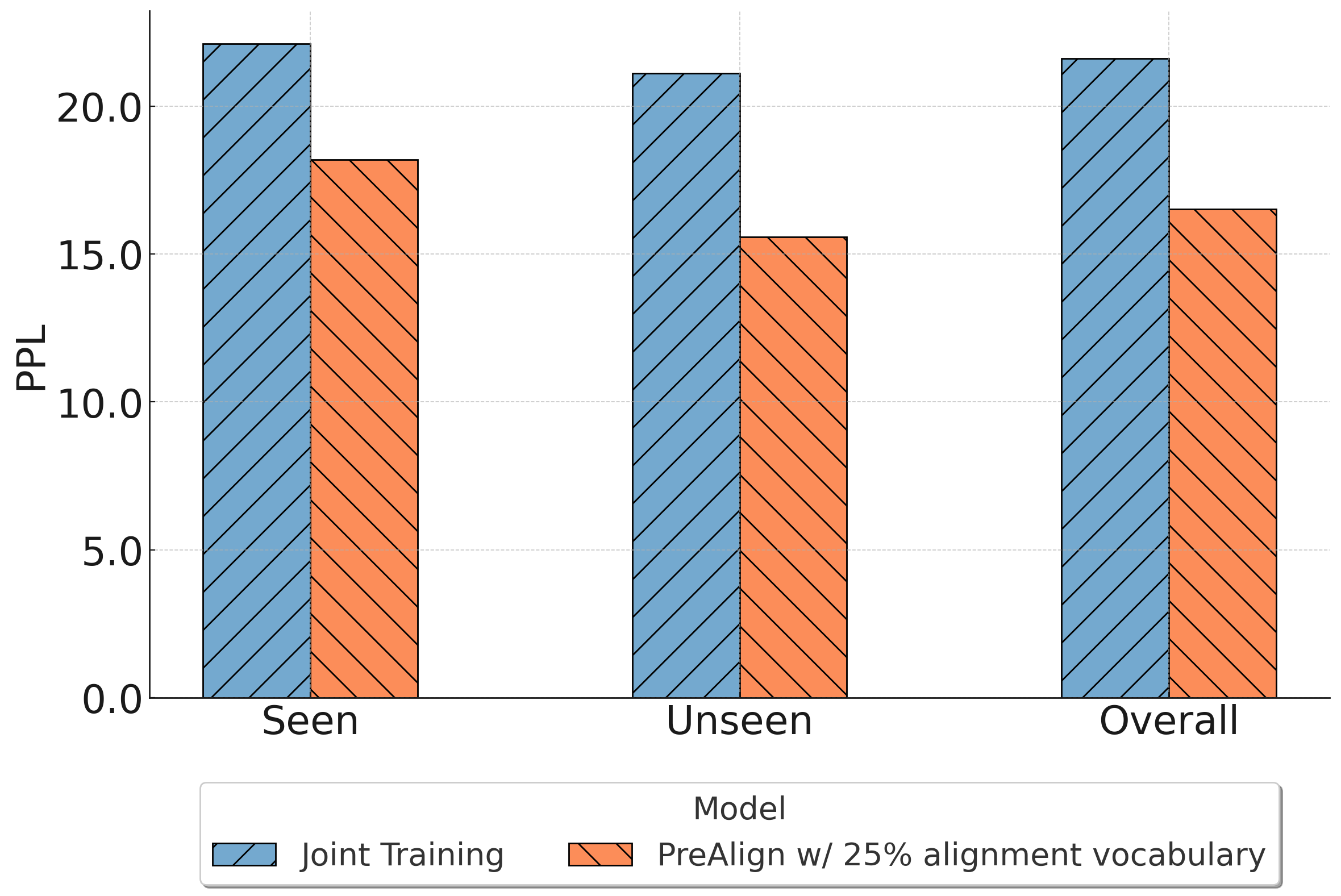}
    \caption{Language modeling perplexity on Seen and Unseen words categorized according to multilingual alignment stage. }
    \label{fig:unseen_ppl}
\end{figure}

 In previous experiments, we assumes that we can collect translations for all words in the pretraining corpus. However, in real-world settings, this might be impractical. Therefore we present an investigation on whether we can only collect alignment table of high-frequency words, and generalize the alignment to words unseen in the alignment table.

Specifically, we sort words in our unique word set according to their frequency, and only train \method model based on the top $\beta$ word alignment. Table~\ref{tab:unseen_portion} shows the results. We can see that when using the most frequent 25\% words for multilingual alignment, \method can already achieve significant improvements over Joint Training. This indicates the alignment information can be generalize between words.

To better validate this, we split all words into Seen and Unseen according to their appearance during the multilingual alignment phase. We then compute the test LM perplexity of seen words and unseen words, and present the results in Figure~\ref{fig:unseen_ppl}. It can be seen that \method not only can effectively leverage seen words to enhance the language modeling ability, but only can generalize the alignment information to unseen words.

\section{Experiments on Real-world Settings}

We validate the effectiveness of \method under real-world settings.  Performances of LM, ZS-CLT and CLKA is shown in Table~\ref{tab:real_world}.

\paragraph{\method are also effective under real-world scenarios.} It can be seen from Table~\ref{tab:real_world} that \method can still achieve substantially better performance compared to the original Joint Training method. This improvements is consistent across different model scales, rendering the effectiveness of \method in real-world scenarios. Interestingly, the transferring effect from English is more preeminent for German and Russian than Chinese and Arabic, indicating typological similarity between language might also play important roles in cross-lingual transferring effectiveness.

\paragraph{Enlarging models is beneficial for CLKA.} We can also see that although Joint Training gets near-random performance at the small scale, the performance grows with the scale of model parameters. This indicates that the ability of spontaneous multilingual alignment only appears on larger models, which is consistent with finding in \citet{qi-etal-2023-cross}.

\section{Conclusion}
We present the \method framework in this paper. It advances the establishment of multilingual alignment prior to language pretraining, and maintain it throughout pretraining using an input-only codeswitching strategy. Through extensive experiments and analysis, both on synthetic and real-world settings, we demonstrate the effectiveness of \method for facilitating cross-lingual ability and knowledge transfer.
\section*{Limitations}
The main limitation of this paper is scale of studied models and datasets. Although we proved the effectiveness of \method up to 1.3B models, it is still very small compared to LLMs nowadays. Whether the findings in the paper holds on larger settings still remains to be explored.

Another limitation is that we only test simple factual knowledge in this paper. In real worlds, knowledge may take more complex forms, and the effectiveness of \method on these settings need to examined.

\section*{Acknowledgement}
We would like to thank the anonymous reviewers for their insightful comments. Shujian Huang is the corresponding author. This work is supported by National Science Foundation of China (No. 62376116, 62176120) and Nanjing University-China Mobile Communications Group Co.,Ltd. Joint Institute.

\bibliography{anthology,custom}

\appendix

\section{Experimental Details}
\label{sec:experimental_details}

\paragraph{Collection of the multilingual table} We recognize words using the \textit{word\_tokenize} function from NLTK library. The word set consists of all words, including named entities, that appears above 20 times in the pretraining corpus. We translate all words using GPT-4, by asking it to generate 5 most common translations of a given word (without placing it in a context).

\paragraph{Training details} During the multilingual alignment stage, we set the $\tau$ to be $0.1$. During the language pretraining stage, we independently sample sentences for LM loss and word pairs for the alignment loss at each training step. We ran all experiments on $8\times$A100 GPUs. The multilingual alignment stage takes about 500 steps, and the language pretraining stage takes about 24000 steps. The running time of different sizes of models ranges from 4 hours to 24 hours.

\paragraph{Example of the synthesized knowledge}
 We collect relations from WikiData, and ask GPT-4 to compose templates for each relation. We then fill in the person name to synthesize knowledge about people. For example, if the subject, relation and object are \textit{Oprah Winfrey, godparent} and \textit{Tyler Perry} respectively, then the composed knowledge is \textit{Oprah Winfrey is the godparent of Tyler Perry.}


\begin{table}[ht]
\footnotesize
\centering
\begin{tabular}{llll}
\toprule
               & En            & De            & Zh            \\ \midrule
\multicolumn{4}{l}{\textbf{150M}}                              \\ \midrule
Joint Training & \textbf{90.6} & 71.5          & 78.9          \\
\method       & 90.1          & \textbf{76.1} & \textbf{83.7} \\ \midrule
\multicolumn{4}{l}{\textbf{400M}}                              \\ \midrule
Joint Training & \textbf{92.6} & 75.1          & 83.3          \\
\method       & 92.3          & \textbf{78.9} & \textbf{85.6} \\ \midrule
\multicolumn{4}{l}{\textbf{1.3B}}                              \\ \midrule
Joint Training & 94.1          & 79.7          & 85.2          \\
\method       & \textbf{94.3} & \textbf{82.4} & \textbf{87.9} \\ \bottomrule
\end{tabular}
\caption{ZS-CLT performance of Joint Training and \method across different scale of models on the PAWSX dataset.}
\label{tab:pawsx}
\end{table}

\section{Experimental results on the PAWSX dataset}
To further validate the effectiveness of PreAlign, we conduct additional experiments on the PAWSX dataset~\citep{yang-etal-2019-paws}, and show the result in Table~\ref{tab:pawsx}.  It can be seen that PreAlign can still bring consistent improvements across different model scales.



\end{document}